%% file: main.tex
\documentclass[10pt, a4paper, twocolumn]{article}

\usepackage{booktabs}
\usepackage{graphicx}
\usepackage{hyperref}
\usepackage{listing}
\usepackage{listings}

\usepackage{float}
\newfloat{listing}{tbp}{lop}
\floatname{listing}{Listing}

\title{Sentiment Analysis of German Sign Language Fairy Tales}

\author{Fabrizio Nunnari, Siddhant Jain, Patrick Gebhard\\
  \texttt{\{fabrizio.nunnari, siddhant.jain, patrick.gebhard\}@dfki.de} \\
  German Research Center for Artificial Intelligence (DFKI) \\
  Saarland Informatics Campus D3 2, 66123 Saarbrücken, Germany
 }

\begin{document}

\maketitle

\begin{abstract}
We present a dataset and a model for sentiment analysis of German sign language (DGS) fairy tales. First, we perform sentiment analysis for three levels of valence (negative, neutral, positive) on German fairy tales text segments using four large language models (LLMs) and majority voting, reaching an inter-annotator agreement of 0.781 Krippendorff’s alpha. Second, we extract face and body motion features from each corresponding DGS video segment using MediaPipe. Finally, we train an explainable model (based on XGBoost) to predict negative, neutral or positive sentiment from video features.
Results show an average balanced accuracy of 0.631. A thorough analysis of the most important features reveal that, in addition to eyebrows and mouth motion on the face, also the motion of hips, elbows, and shoulders considerably contribute in the discrimination of the conveyed sentiment, indicating an equal importance of face and body for sentiment communication in sign language.\\
\textbf{Keywords:} German sign language, DGS, sentiment analysis, fairy tales, explainable AI, machine learning.
\end{abstract}

\section{Introduction}

Sign languages (SLs) are the native language of 70 million people in the world \cite{eberhard_ethnologue_2025} and they are the main communication languages among Deaf communities. SLs rely on hands, fingers, torso, shoulders, gaze, head motion and facial expressions to convey meaning.

Although performance of SL translation pipelines is still rudimentary, some companies start offering online automated translation tools (e.g.: Signapse - \url{https://www.signapse.ai}, Migam - \url{https://migam.ai}, and Nagish - \url{https://sign.mt}).

The output of SL synthesis systems is however criticized as being too robotic and unexpressive, to the extent that the semantics of the message is not correctly communicated.
One reason for that, is the lack of explicit support for the communication of ``sentiments'', and ``emotions'' in SL translation systems.

Expressing emotions in spoken communication (e.g., through facial expressions, hand gestures, and body postures) can be seen as an addition to the basic message conveyed via words. In contrast, in SL the whole body and face are already involved in the communication of the semantics of the message, and such a dichotomy between grammatical communication and emotion expression is not possible.

Research on emotion expression in SL is still emerging (e.g., see the ExEmSiLa workshop \cite{herrmann_expressing_2024}), but it is carried on mainly by SL linguists through time-consuming manual annotation and human observations of video material, without the support of modern automated machine learning tools.

In contrast, for written languages, sentiment analysis is a well established branch of natural language processing (NLP) that automatizes the inference of an emotion (e.g., happiness, sadness) or the valence (positive, neutral, negative).

This paper investigates the automated sentiment analysis of SL.
The goal being to deploy a computational model that can correctly infer valence from the analysis of SL video segments\footnote{We refer to a SL video clip conveying a semantic information as a \emph{segment}, instead of sentence, because there is no exact correspondence between the two.}.
Valence (also known as Pleasure) is one of the three dimensions used to quantify emotions in the PAD space (Pleasure/Valence, Arousal, Dominance) \cite{mehrabian_pleasure-arousal-dominance_1996}. In this first investigation, we target only the valence dimension. 

\begin{figure}
    \centering
    \includegraphics[width=1.0\linewidth]{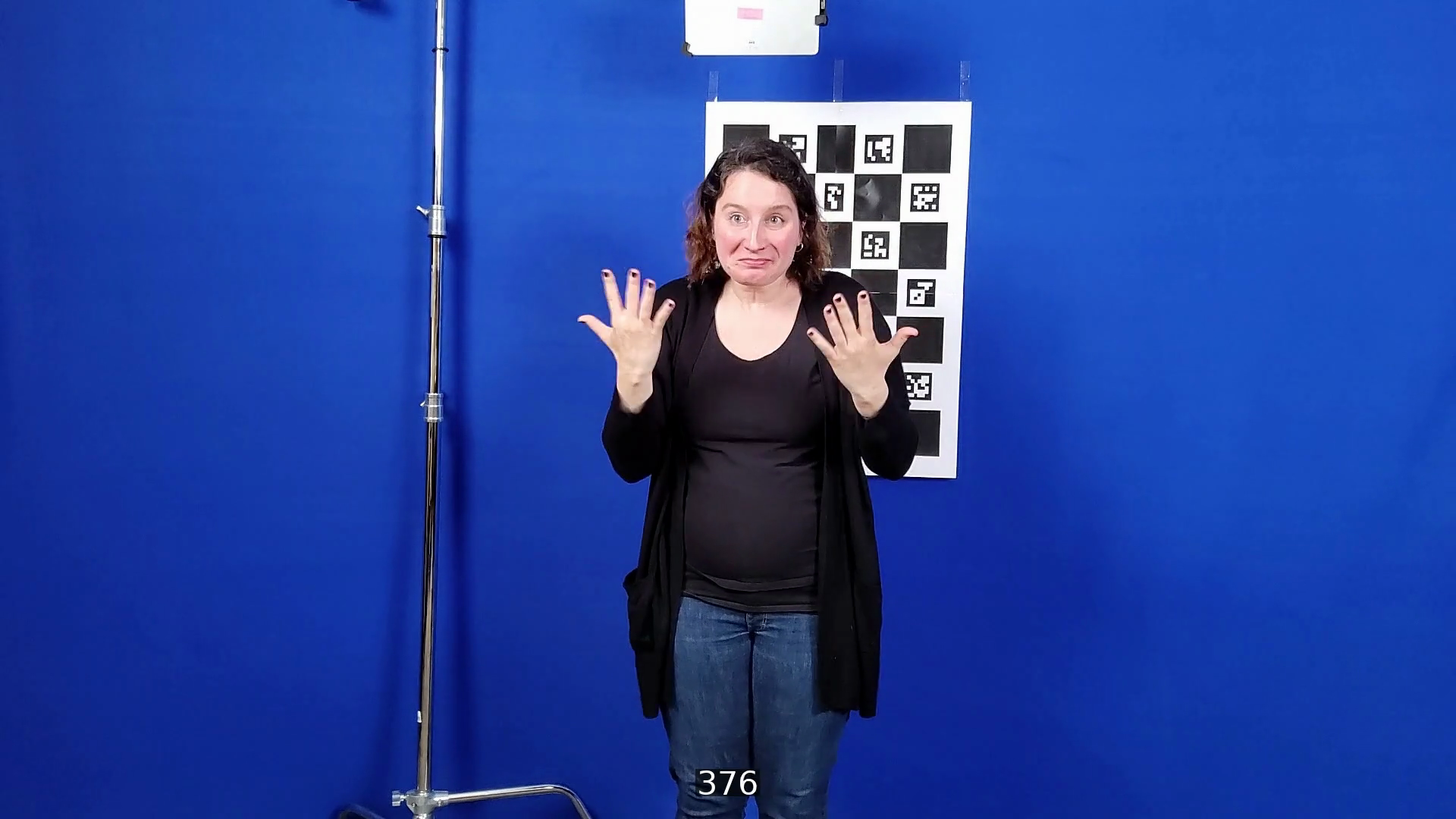}
    \caption{A frame from the first sentence of the tale \emph{The Hare and the Hedgehog} from the DGS-Fabeln-1 corpus. The corresponding German sentence is ``Der Hase und der Igel. Es war einmal: So fangen Märchen an. Ein Märchen ist eine sehr alte Geschichte. Dieses Märchen heißt: Der Hase und der Igel. Das Märchen geht so:'' [ENG: ``The Hare and the Hedgehog. Once upon a time: that’s how fairy tales begin. A fairy tale is a very old story. This fairy tale is called: The Hare and the Hedgehog. The fairy tale goes like this:'']. The corresponding video lasts about 15 seconds.}
    \label{fig:dgs-fabeln-1-example}
\end{figure}

For this work, we rely on the DGS-Fabeln-1 dataset \cite{nunnari_dgs-fabeln-1_2024,nunnari_dgs-fabeln-1_2024-1}, a parallel corpus of 7 German sign language fairy tales, consisting of 574 German text segments each associated with a video clip.
Figure \ref{fig:dgs-fabeln-1-example} shows an example.

We focus on fairy tales because they naturally cover the whole range of positive to negative valence, thus displaying a high degree of expressivity in sign language performance.
This seems to be a better candidate with respect to the most popular dataset for DGS used in natural language processing, i.e., the RWTH-PHOENIX-2014-T corpus \cite{camgoz_neural_2018}, which is limited to the domain of weather forecast.

To perform our tests, we first take advantage of the state-of-the-art performance of LLMs in sentiment classification \cite{klahn_dictionaries_2025}. Four publicly available large language models (LLMs) are used to extract sentiment labels from the text segments, and a final label is assigned through majority voting. Second, the MediaPipe library \cite{lugaresi_mediapipe_2019} is used to analyse SL videos and extract blend-shapes and landmarks, which are further elaborated to extract information about velocities, accelerations, mutual distances, accumulated distances, bone rotations, and peak frequency. Finally, XGBoost is used to build a model predicting sentiment from video features. The details are reported in section~\ref{sec:method}.

This paper brings the following three contributions. 
\textbf{First}, we release DGS-Fabeln-1-SE (Sentiment Estimation), which extends DGS-Fabeln-1 with: i) the sentimental labels predicted by the 4 LLMs and their majority voting, and ii) the MediaPipe motion data of all segments, including an extra set of features precomputed for the classification task. \textbf{Second}, we deploy an explainable prediction model, based on XGBoost, capable of predicting sentiment in DGS-Fabeln-1 video segments.
\textbf{Third}, we report and comment on the list of the most important features for the classification of valence from SL videos.

The DGS-Fabeln-1-SE dataset is publicly available at \url{https://doi.org/10.5281/zenodo.18879038}. The code to process the dataset is available at \url{https://github.com/DFKI-SignLanguage/LREC2026-DGS-Fabeln-1-SE}

\section{Related work}

\subsection{Sentiment analysis through LLMs}

Manual sentiment annotation is a time- and resource-consuming task. Given the increased reliability of automated sentiment analysis, we opted for a procedural  annotation of the DGS-Fabeln-1 dataset.

The system introduced by \cite{guhr_training_2020}
is considered among the best for sentiment analysis on the German language. However, an early test on fairy tales showed a peculiar behavior, with more than 90\% of the text chunks classified as ``neutral''. A manual check on a few samples led us to consider such library inappropriate for this task.

Indeed, Klähn et al. \cite{klahn_dictionaries_2025} report that while fine-tuned German BERT models achieve 94-96\% F1 scores on contemporary German text,
these models experience catastrophic performance degradation in other contexts, like historical German. In contrast, LLMs can generalize much better in unseen domains (zero-shot learning).
For example, also Suter and Mecker \cite{suter_using_2024} report that GPT 4 demonstrated a ``substantial to near perfect agreement'' with human annotators on German text classification.

In our work, we relied on a combination of the latest available versions of four different LLMs (GPT5, Sonic, Mistral, GPT OSS 20B) and ran a majority voting to compensate for disagreements.

\subsection{Sentiment Analysis on SLs}

\cite{oguike_using_2025} performed sentiment analysis on Indian sign language. 
They trained a mixed model using both video and text to predict labels in 4193 frames, singularly annotated for negative, neutral, and positive valence. They claim that, using VGG16, the video-only modality reaches $>$99\% accuracy.
In contrast, we performed the analysis on full sentences, rather than single frames, and used an explainable model listing the human-understandable features that contribute to the classification.

\cite{takir_sentiment_2025} investigated a 3-level sentiment analysis for single signs of the Turkish sign language (TİD). They used a late fusion architecture merging the classification based on face and hands separately, reporting that the hands-only modality reaches the worst accuracy (42\%) with respect to using face-only (46\%) or their combination (65\%).
The main difference from our work is that they perform the classification on short single-sign videos, while we classify full sentences.

\cite{wang_towards_2021} implemented a system that converts sign language into emotionally driven speech. To achieve it, they first analyse facial expressions from SL videos and use the derived pleasure-arousal-dominance (PAD) values to drive the speech synthesis. Results show a good match between the emotions recognized in the facial expressions w.r.t. speech, but this doesn't assess the validity of using PAD recognition as a good indicator of emotion in SL, as the face is simultaneously involved in performing grammar roles.

\cite{zhang_u-shaped_2024} built a neural model tested on a dataset of 50 sentences in Chinese sign language from 18 participants under 5 emotions. The emotions were derived from a clusterization of 5 areas of the U-shape typically formed by correlating Valence and Arousal classification on sentiments. They claim 88\% accuracy in the classification task. Their system, based on wearable armband with built-in surface electromyograph (sEMG) and inertial measurement unit (IMU) sensors is unpractical for every-day use, but their results confirm that a significant part of the emotions of a message is carried by body movement.

\cite{chua_emosign_2025} introduced recently a dataset of 200 sentences for sentiment analysis on American sign language (ASL). Each video was annotated by three SL experts for sentiment valence, 10 emotions, and free comments.
The creation process represents the state-of-the-art in the field, but the resulting annotations present an averge inter-annotator agreement on only 0.53 in terms of Krippendorff’s Alpha, with a couple of labels scoring less than 0.2. This highlights the difficulty and subjectivity of human annotation for SL videos.
Differently, in our work we use automated annotation only on the textual side of the corpus: a task for which LLMs show a satisfying accuracy and higher agreement.

\cite{chua_emosign_2025} also presented an attempt to automatize the prediction of emotions from videos. They based their work on multimodal LLMs with video analysis capabilities, trying to predict emotion labels from an encoding of the video downsampled at 10Hz rate. The prediction of 3-levels of sentiment reaches 56.18\% accuracy using AffectGPT, but with the need of high computation power and without an explanation of the result.
In contrast, in this work we employ a light-weight preprocessing phase for body/face motion features extraction, followed by a very light-weight XGBoost predictor (possibly enabling real-time inference) which gives by-product a list of the most relevant features.

\subsection{Sentiment Analysis on Fairy Tales}

There are several reasons on why to prefer fairy tales (over other domains such as weather forecast) to perform sentiment analysis.

First, fairy tales display much larger standard deviations across all emotion categories, indicating greater variability in emotional content within individual tales \cite{mohammad_once_2011}.


\cite{volkova_emotional_2010} report that fairy tales present distinctive challenges and opportunities for sentiment analysis. The genre’s formulaic nature, with clear archetypal characters and predictable narrative structures, provides well-defined emotional contexts that facilitate automatic sentiment detection.

In addition, \cite{volkova_emotional_2010} have identified specific emotional trajectory patterns in fairy tales, with polynomial fitting revealing wave-shaped emotional flows common across different tales.
Coherently, \cite{mohammad_once_2011} reports that the emotional dynamics follow predictable patterns from initial states through conflict development to resolution, often culminating in positive outcomes that justify the “happily ever after” convention.
We will use those observations to verify the plausibility of LLM-base automated sentiment prediction on our fairy tales text.

Finally, it seems that the high emotional variability within individual tales suggests complex emotional journeys that engage readers through dynamic sentiment shifts rather than static positive messaging \cite{mohammad_once_2011,herrmann_fairy_2023}.
This is in accordance to our first observations that valence may change rapidly even within a single video segment. For this reason, as described later in the method section, we implemented a procedure to detect and exclude segments with mixed emotions.



%
%
%
\section{Method}
\label{sec:method}

\subsection{Data: DGS-Fabeln-1}

For our data preparation, we use the DGS-Fabeln-1 \cite{nunnari_dgs-fabeln-1_2024,nunnari_dgs-fabeln-1_2024-1}, a dataset of seven German fairy tales interpreted in German sign language (DGS) from simplified German text.
Each fairy tale is divided into text segments (between 37 and 150) for a total of 574 segments. Each segment, roughly corresponding to 2.5 sentences, is associated with a video clip. Video clips show the DGS interpretation of the corresponding text, each lasting 9.6 seconds on average.
The dataset totals 92 minutes of video recording.
Figure \ref{fig:dgs-fabeln-1-example} shows an example.

\subsection{Method overview}


\begin{figure*}
    \centering
    \includegraphics[width=1.0\textwidth]{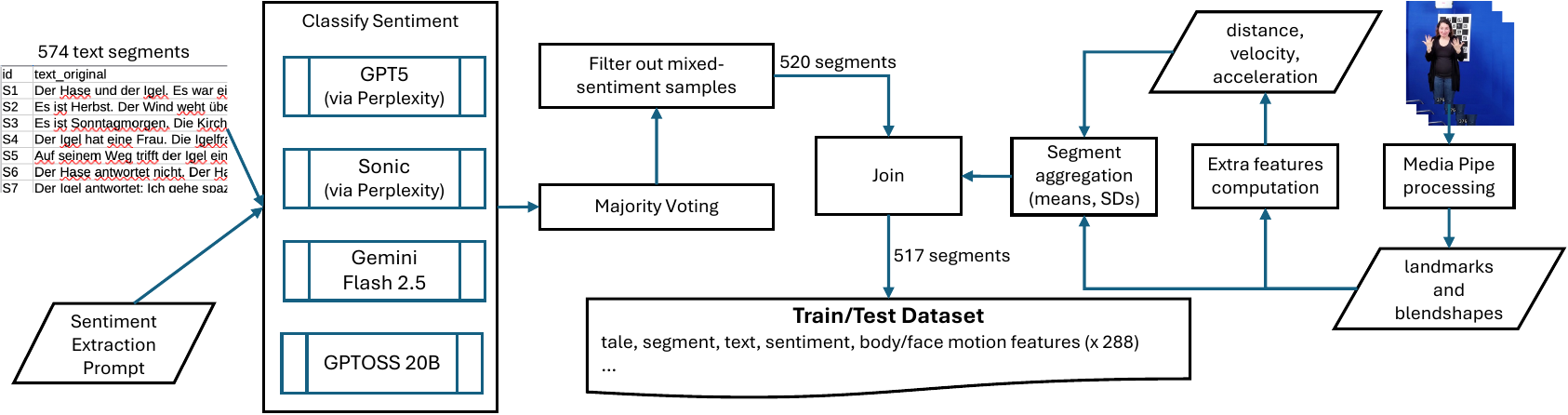}
    \caption{Overview of our dataset preparation pipeline.}
    \label{fig:dataset-preparation}
\end{figure*}

Figure \ref{fig:dataset-preparation} shows the steps for the preparation of the training/testing data.

As depicted on the left side, the text segments of each fairy tale were given as input to four LLMs, encapsulated in a prompt asking to perform sentiment analysis separately on each segment, and also indicating whether more sentiments are applicable.
One single label per segment is then chosen via majority voting. Text segments presenting mixed emotions or no agreement were discarded.

On the right side, Figure \ref{fig:dataset-preparation} depicts the video processing steps.
Each video segment is processed with the MediaPipe library \cite{lugaresi_mediapipe_2019}.
The raw 3D landmark locations and facial blendshapes are used to compute an extra set of manually engineered features, and then aggregated at segment level via means and standard deviations.

Predicted sentiments and video aggregated features are then joined into the DGS-Fabeln-1-SE dataset. Because of the absence of some videos in the original DGS-Fabeln-1 corpus, only 517 segments are finally available.

\subsection{Feature extraction from videos}

Each frontal video segment from the DGS-Fabeln-1 dataset was processed using a multimodal feature extraction pipeline based on MediaPipe Holistic and Face Landmarker v2 models \cite{lugaresi_mediapipe_2019}. The aim was to obtain consistent three-dimensional trajectories of upper body, hands, head, and facial blendshapes for every frame.
The following landmark sets were extracted:
    \emph{Pose landmarks} for wrists, torso, shoulders, elbows, hips, and nose;
    \emph{Facial blendshapes and transformation matrix} for 52 blendshape activations and a 4×4 affine transform containing head rotation and translation.

Because landmark detection occasionally fails (e.g., during occlusion or motion blur), missing values were linearly interpolated over time to ensure frame continuity.

From the interpolated time series, additional features, commonly used in gesture recognition, were computed to better describe the motion dynamics.
\begin{itemize}
    \item \textbf{Velocity and acceleration} for wrists, shoulders, hips, and nose.
    \item \textbf{Euclidean distances} between: i) the two wrists, ii) the two elbows iii) a wrist and its corresponding shoulder, iv) a wrist and the opposite shoulder, v) wrists and the nose.
    \item \textbf{Head pitch, yaw, and roll} angles computed with a decomposition of the 3×3 rotation submatrix of the head transform.
    \item \textbf{Left and right arm elevation angles} computed from the shoulder–elbow vectors relative to the torso axis.

\end{itemize}

For each video, all of those frame-level features were aggregated into the following statistics:
\begin{itemize}
    \item \textbf{Mean}.
    \item \textbf{Standard deviation}.
    \item \textbf{Accumulated movement distance} for wrists, shoulders, and nose.
    \item \textbf{Peaks per second} calculated on each raw data: useful to catch signal sparks that might be hidden by average aggregations over long sequences. Peaks were computed via the \texttt{find\_peak} function of the SpiPy framework\footnote{\url{https://docs.scipy.org/doc/scipy/reference/generated/scipy.signal.find_peaks.html}} and then normalized by video duration.
\end{itemize}

This yielded to a feature vector of 396 items per video segment.

\subsection{Sentiment extraction from text}

The sentiment analysis was performed by aggregating the votes of four large language models: GPT5, Sonic, Gemini Flash 2.5, and GPT OSS 20B.
For GPT5 and Sonic, the analysis was performed through the Perplexity Pro app. Gemini was used through its web interface. Finally, GPT OSS 20B was run on a local machine using the ollama\footnote{\url{https://ollama.com}} service and its Python API.

All sentiment voting was performed using the prompt reported in Listing \ref{lst:sentiment-task}, followed by the tale sentences, each surrounded by double quotes and separated by an extra space.

\begin{listing*}[t]
\begin{lstlisting}[caption={Sentiment analysis prompt}, label={lst:sentiment-task}, basicstyle=\scriptsize\ttfamily]
Consider the task of Sentiment analysis (also called opinion mining)
in the field of Natural Language Processing (NLP).
I will give you a list of German text chunks.
For each chunk, evaluate its "sentiment" on a 3 levels scale: negative, neutral, positive.
Also, for each chunk, mark if there is more than one sentiment that can be applied
(for example, if a long text chunk changes sentiment),
and report the sentiments separated by a dash (-)

Format the output as a three-column comma-separated values (CSV), using the comma as separator.
The first column will contain the original text chunk.
The second column will contain the (list of) sentiment(s) that you evaluated.
The third column will contain "yes" if multiple sentiments were found, "no" otherwise.
Also, output a header with column names: "Text", "Sentiments", "Multi".

As a consequence, the number of output rows must be the same as the number of input text chunks.

In the output, omit any kind of preamble and explanation. Output only the CSV data.
Do not use double quotes in the output CSV.

Here is the list of chunks:
\end{lstlisting}
\end{listing*}

The goal of the prompt is to perform the sentiment judgment and to spot if mixed sentiments appear in the same text chunk. Our goal was to skip such segments in order to avoid the treatment of a multi-label problem.

The resulting votes were then aggregated into a single column through majority voting. The majority agreement on a segment is reached if a label has more counts than any other. However, a segment was marked as ``no-agreement'' if either:
\begin{itemize}
    \item Two (or more) labels have the same count, or
    \item the majority label is itself a mixed emotion.
\end{itemize}

%
\begin{figure}[b]
    \centering
    \includegraphics[width=0.9\linewidth]{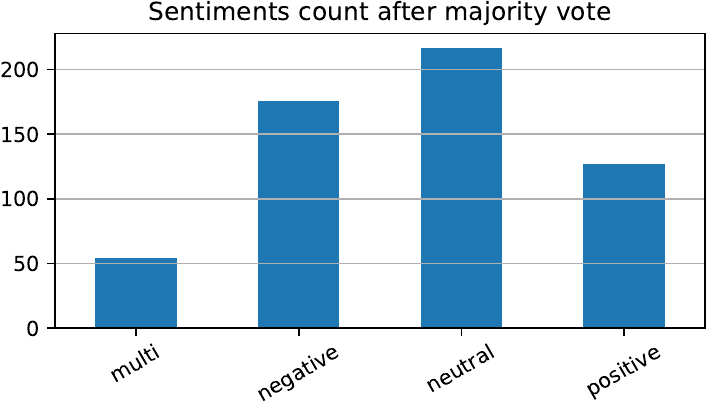}
    \caption{Distribution of sentiment among the 574 sentences of our dataset.}
    \label{fig:sentiment-label-distribution}
\end{figure}

Figure \ref{fig:sentiment-label-distribution} shows the final distribution of the labels after aggregation.
In total, 520 segments passed the majority voting filter.


As a measure of inter-annotator agreement, we computed the Krippendorff's alpha coefficient \cite{krippendorff_content_2019}, using the Castro's implementation for Python
\cite{castro_fast_2017}.
Table \ref{tab:agreement-alpha} shows the alpha values of the collected votes before and after applying the majority voting filter. As expected, the alpha coefficient increases (from $0.715$ to $0.786$) after filtering out the items with no majority applicable.

\begin{table}[h]
    \centering
    \scriptsize
    \begin{tabular}{c|cc}
    \toprule
      Data & Samples & Krippendorff's alpha \\
    \midrule
      Raw votes & 574 & $0.715$ \\
      After majority voting & 520 & $0.786$ \\
    \bottomrule
    \end{tabular}
    \caption{Inter-annotator agreement among the four LLMs used for sentiment analysis}
    \label{tab:agreement-alpha}
\end{table}

%

\begin{figure*}
    \centering
    \includegraphics[width=0.245\textwidth]{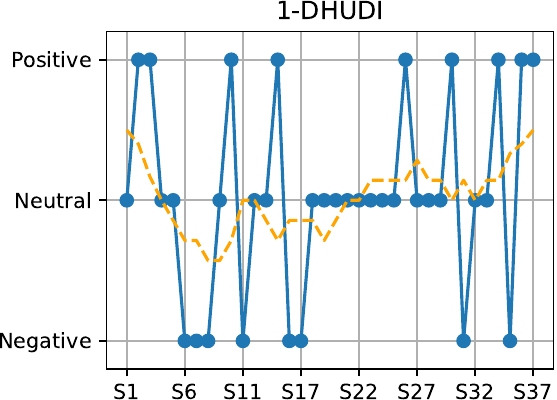}
    \includegraphics[width=0.245\textwidth]{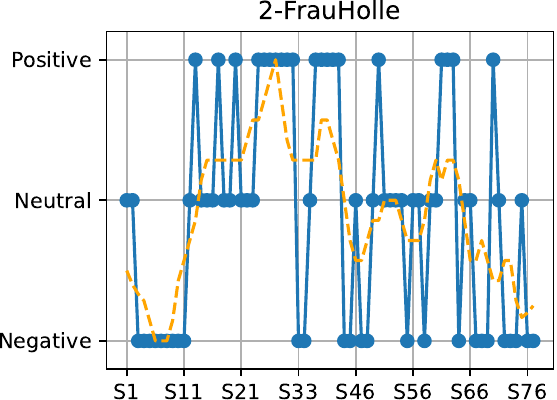}
    \includegraphics[width=0.245\textwidth]{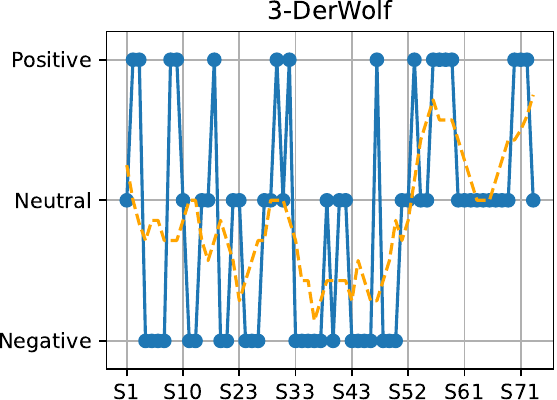}
    \includegraphics[width=0.245\textwidth]{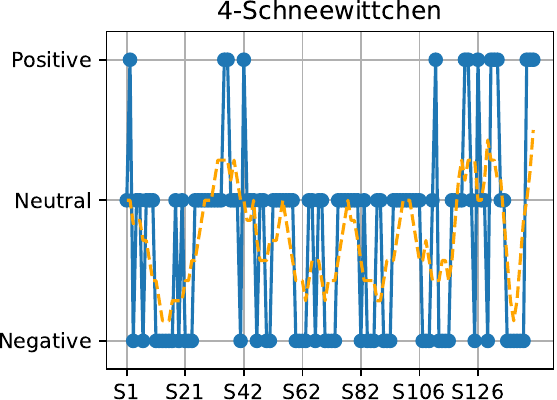}\\
    \includegraphics[width=0.25\textwidth]{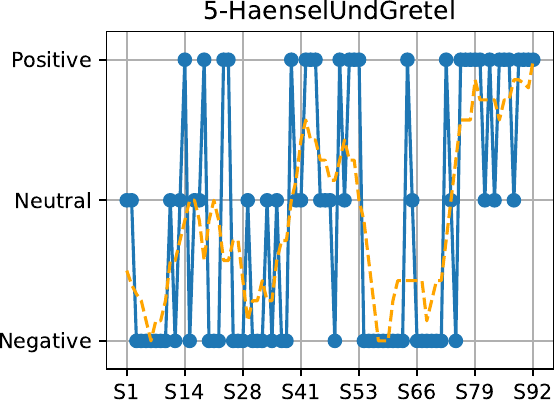}
    \includegraphics[width=0.25\textwidth]{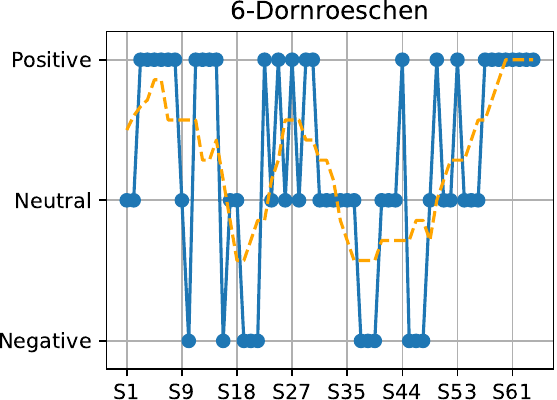}
    \includegraphics[width=0.25\textwidth]{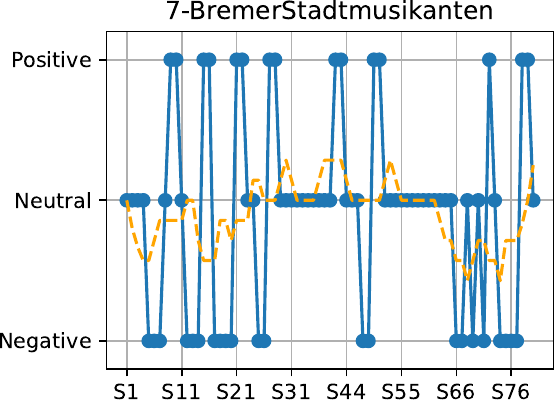}
    \caption{For each fairy tale, the sentiment is plotted against the evolution of the plot. The dashed line shows the rolling mean on a 7-sample moving window.}
    \label{fig:sentiment-evolution}
\end{figure*}


As a further step to qualitatively assess the validity of sentiment votes, we plotted, for each tale, how the sentiment evolves during the narration (Figure~\ref{fig:sentiment-evolution}).
It is possible to observe, coherently with the patterns reported in the literature, how the evolution of the sentiment follows a trajectory of alternating states \cite{volkova_emotional_2010} which (with the exception of Frau Holle) culminates into a positive ``happy end''~\cite{mohammad_once_2011}.

\subsection{Evaluation Metrics}

For each cross-validation fold, predictions were evaluated in terms of overall accuracy, balanced accuracy, precision, recall, and F1-score. Balanced accuracy, which is defined as the mean of per-class recall values, was used to mitigate bias toward more frequent sentiment categories.  

Precision, recall, and F1-scores were computed in both macro and weighted forms. Macro-averaged metrics assign equal weight to each class, reflecting the model’s ability to generalize across sentiments, while weighted averages account for class frequency for overall stability across folds.

Finally, for a comparison with other sentiment classification tasks, we computed a Pearson $\rho$ correlation after mapping negative, neutral, and positive labels to $-1$, $0$, and $+1$, respectively.

All metrics were calculated using the scikit-learn library \cite{pedregosa_scikit-learn_2011}. The macro-averaged F1-score was used as the primary criterion for model selection during hyperparameter optimization.

\subsection{Training}

Our work relies on the XGBoost classifier \cite{chen_xgboost_2016}.
The model preparation was conducted in two stages to ensure robust generalization across tales.
In the first stage, we performed a grid search for the best hyper-parameters applying a partitioning into folds via a \textbf{Stratified Group K-Fold (SGKF)}\footnote{\url{https://scikit-learn.org/stable/modules/generated/sklearn.model_selection.StratifiedGroupKFold.html}} approach, using macro-F1 as the selection criterion, with early stopping after 150 rounds, grouping by \textit{Tale} to prevent data leakage between segments originating from the same instance. Stratification maintained the sentiment class distribution in each fold.

The grid search was performed over a compact set of hyperparameters controlling model complexity, learning rate, and regularization: \emph{max\_depth}, \emph{min\_child\_weight}, \emph{eta}, \emph{gamma}, \emph{subsample}, \emph{colsample\_bytree}, \emph{lambda}, \emph{alpha}, \emph{scale\_pos\_weight}. Table \ref{tab:hyper-parameters} reports the selected hyper-parameters values.

\begin{table}
  \centering
  \scriptsize
  \caption{Selected XGBoost hyper-parameters.}
\begin{tabular}{ll}
\toprule
Parameter & Selected Value \\
\midrule
\texttt{max\_depth} & $5$ \\
\texttt{min\_child\_weight} & $1.5$ \\
\texttt{eta} & $0.06$ \\
\texttt{gamma} & $0.15$ \\
\texttt{subsample} & $0.85$ \\
\texttt{colsample\_bytree} & $0.75$ \\
\texttt{lambda} & $2.0$ \\
\texttt{alpha} & $0.6$ \\
\texttt{scale\_pos\_weight} & $0.9$ \\
\bottomrule
\end{tabular}
\label{tab:hyper-parameters}%
\end{table}


In the second stage, we trained the model using a standard \textbf{five-fold stratified cross-validation} to obtain the final performance estimates.


\subsection{Feature selection}

To discard confounding features and improve model accuracy, we sorted the prediction features according to their mean importance across fold. We then repeated training and test using only the top-N most important features (selecting groups of 16, 32, 64, 96, 128, and 160). The best performance was achieved using the first 96 features.

\section{Results}

\subsection{Performance}

\begin{figure}[t]
    \centering
    \includegraphics[width=1.0\linewidth]{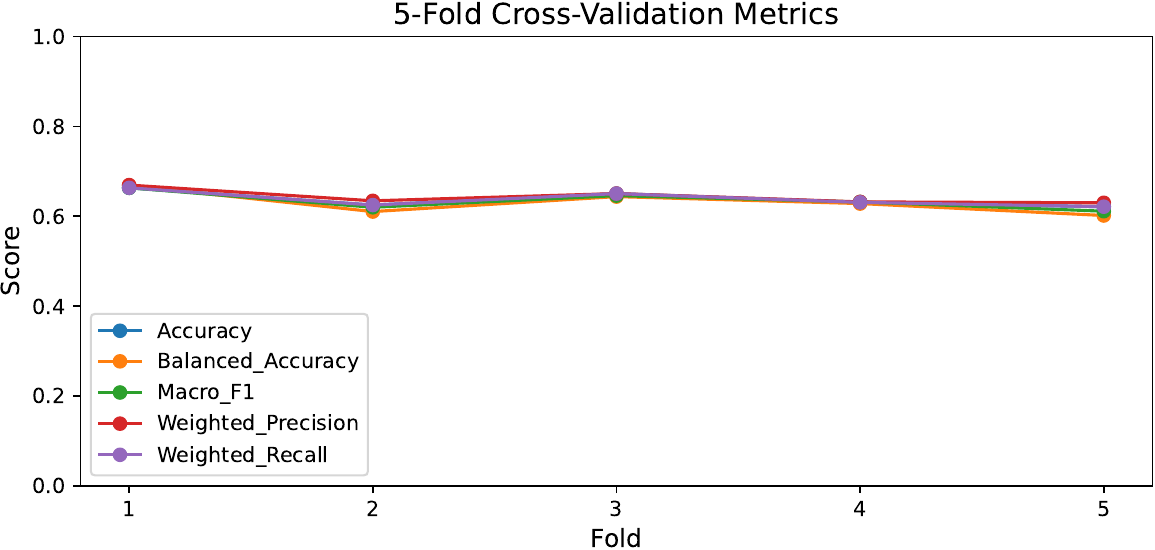}
    \caption{Test metrics per folder.}
    \label{fig:cross-validation-metrics}
\end{figure}

\begin{table*}[t] 
  \centering
  \scriptsize
  \caption{Metrics computed on the 5 folds and their average.}

\begin{tabular}{l|rr|rrr|rrr|rrr|r}
\multicolumn{1}{r}{} &       &  \textbf{Bal}.   & \multicolumn{3}{c|}{\textbf{Macro}} & \multicolumn{3}{c|}{\textbf{Weighted}} & \multicolumn{3}{c|}{\textbf{Recall}} &  \\
\textbf{Fold} & \multicolumn{1}{l}{\textbf{Acc.}} & \multicolumn{1}{l|}{\textbf{Acc.}} & \multicolumn{1}{l}{\textbf{Prec.}} & \multicolumn{1}{l}{\textbf{Recall}} & \multicolumn{1}{l|}{\textbf{F1}} & \multicolumn{1}{l}{\textbf{Prec.}} & \multicolumn{1}{l}{\textbf{Recall}} & \multicolumn{1}{l|}{\textbf{F1}} & \multicolumn{1}{l}{\textbf{Neg.}} & \multicolumn{1}{l}{\textbf{Neut.}} & \multicolumn{1}{l|}{\textbf{Pos.}} & \multicolumn{1}{l}{\textbf{Pearson $\rho$}} \\
\midrule
\multicolumn{1}{r|}{1} & 0.663 & 0.669 & 0.663 & 0.669 & 0.663 & 0.670 & 0.663 & 0.664 & 0.686 & 0.628 & 0.692 & 0.509 \\
\multicolumn{1}{r|}{2} & 0.625 & 0.611 & 0.645 & 0.611 & 0.620 & 0.634 & 0.625 & 0.623 & 0.657 & 0.674 & 0.500 & 0.527 \\
\multicolumn{1}{r|}{3} & 0.650 & 0.644 & 0.649 & 0.644 & 0.646 & 0.651 & 0.650 & 0.650 & 0.657 & 0.674 & 0.600 & 0.535 \\
\multicolumn{1}{r|}{4} & 0.631 & 0.628 & 0.636 & 0.628 & 0.632 & 0.632 & 0.631 & 0.631 & 0.657 & 0.628 & 0.600 & 0.568 \\
\multicolumn{1}{r|}{5} & 0.621 & 0.602 & 0.637 & 0.602 & 0.611 & 0.630 & 0.621 & 0.619 & 0.611 & 0.714 & 0.480 & 0.507 \\
\midrule
\textbf{Mean} & 0.638 & 0.631 & 0.646 & 0.631 & 0.635 & 0.643 & 0.638 & 0.638 & 0.654 & 0.664 & 0.574 & 0.529 \\
\end{tabular}%

  \label{tab:5-fold-results}%
\end{table*}%

Figure \ref{fig:cross-validation-metrics} shows the resulting main metrics across the five fold. Table \ref{tab:5-fold-results} reports the numerical details.

\begin{figure}[t]
    \centering
    \includegraphics[width=1.\columnwidth]{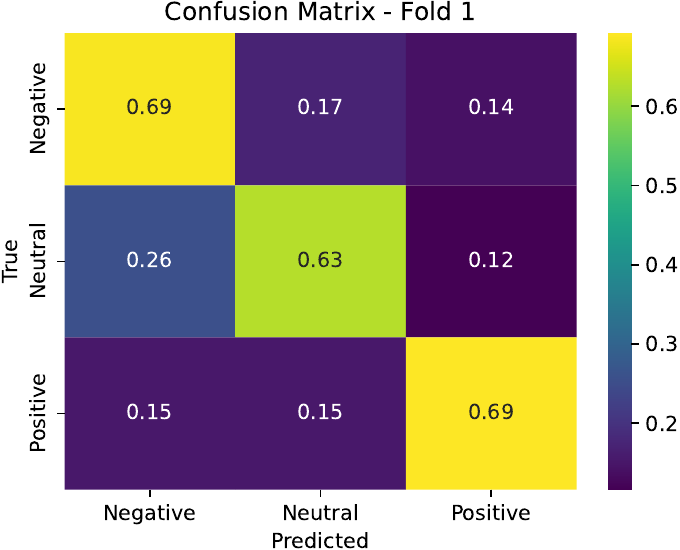}
    \caption{Confusion matrix for test fold 1.}
    \label{fig:conf-matrix-fold1}
\end{figure}

On average, our model reaches 63.1\% balanced accuracy and 63.5\% macro-F1.
Figure \ref{fig:conf-matrix-fold1} shows the confusion matrix for fold 1, the one with the highest accuracy, where it is visible that the most problematic case was for neutral labels predicted as negative.

\subsection{Baseline comparison}

Given the seminal status of the research in sentiment analysis for sign language, we could not identify any suitable baseline for a direct comparison on sign language videos analysis.
However, early work on sentiment prediction on text reported an overall Pearson correlation $\rho=0.65$ \cite{preotiuc-pietro_modelling_2016,buechel_sven_emotion_2016}. In our experiments, after converting classification labels to regression values, we computed a $\rho=0.529$, indicating a lower performance with respect to the text analysis domain.

\subsection{Most important features}

\begin{figure*}
    \centering
    \includegraphics[width=0.9\textwidth]{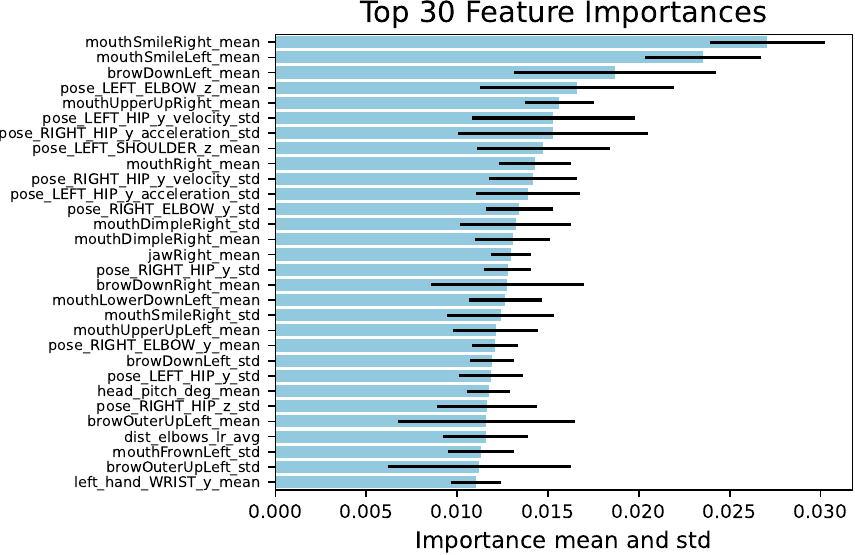}
    \caption{Top 30 features for the prediction of the sentiment labels. Mean and standard deviation among the five folds.}
    \label{fig:top30-features}
\end{figure*}

Figure \ref{fig:top30-features} reports the 30 most important features selected by the model training phase. In the following, we try to give some interpretation of their importance in the prediction of valence.

\textbf{Face smile and eyebrows.} The contribution of the face in communicating emotions in SL has already been investigated, and thus the role of smile, eyebrows (raising or furrowing) or mouth corners (to smile or frown) is not surprising.


\textbf{Left elbow and shoulder distance from the camera.}
According to the mean depth coordinates of the left elbow and shoulder, the sentiment is more positive when the interpreter turns to her right.  This might be explained by the fact that the interpreter often enacted negative sentences while impersonating an evil character, and thus rotating the torso to perform a role-shift. If true, this also suggests a tendency to place the novel's antagonist always on the same side.

The \textbf{mean and standard deviation of the height of the right elbow} have a positive correlation with valence, meaning that more positive sentences are characterized by higher and wider vertical movements of the right elbow (dominant).

\textbf{Hips vertical motion.} There is a negative correlation between  the valence and the standard deviation of the position, velocity, and acceleration of the hips on the vertical axis. This means that for negative sentences the interpreter is performing more variations of motion on her vertical axis. Indeed, it can be noticed from the video performances how she is ``jumping'' on place when the context of the story gets troublesome.

The \textbf{mean distance between elbows} has a positive correlation with valence. Suggesting that more positive utterances are performed by widening the movement of the arms.

An exhaustive analysis and explanation of the correlations is beyond the scope of this work. For future linguistic analyses, the complete list of the 96 features selected for the best performing model is available in appendix~\ref{app:all-features}.

\section{Limitations}

One of the main limitations of this work is that no human annotators participated in the labeling task of the dataset. Although LLMs are recognized for their relatively high performance in sentiment analysis tasks, the contribution of human annotators would set a more reliable and validated ground truth.
Furthermore, given that the main goal of the study is to predict the sentiment of SL videos, the most appropriate setup would include the annotation and agreement performed directly on the video material by native DGS annotators.


As stated earlier, the signer is performing many role shifts, thus biasing the recognition of emotions with the recognition of role taking of a character potentially associate to positive or negative communications.
For an unbiased extraction of the significant features involved in the sentiment prediction, torso rotation should be compensated for by data normalization, or the stimuli material should be selected to prevent role taking.

\section{Conclusions and Future Work}

We presented a method for the construction of a corpus for sentiment analysis on German sign language fairy tales using LLMs, followed by the pipeline for training an explainable prediction model for the inference of sentiment valence from videos.
This is to our knowledge the first work performing a systematic analysis of both face and body contributions for sentiment expression using explainable models.

The results show the ability to predict three-level sentiment valence with a mean balanced accuracy of 63.1\% and a macro-F1 score of 63.5\%. About half of the most important features for the prediction belong to the body, which suggests the importance of joint face and body analysis for the prediction of valence from sign language utterances.\\

Our research could be improved in several ways.
First, our analyses run on full video segments, lasting almost 10 seconds on average. However, sentiment manifestation is often expressed by specific key signs, whose contribution might be absorbed within such long video sequences. To partially address this issue, we introduced only the ``peak count'' features, while other approaches may be available. An alternative approach would employ an continuous analysis on shorter moving time windows.

Second, for this study, we used only the frontal videos of the DGS-Fabeln-1 using the MediaPipe library, which is popular for its speed but less for its accuracy. Better body/face motion analysis could be achieved by using more accurate (although slower) systems (such as OpenPose: \url{https://github.com/CMU-Perceptual-Computing-Lab/openpose}), or by simultaneously using the seven viewpoints available in the DGS-Fabeln-1 corpus, which would allow a triangulation for better landmarks and face blendshape estimation.

A more comprehensive discussion about the reasons why some of the features are so relevant for the recognition of sentiment valence is beyond the scope of this work. However, we believe that our presented findings can help researchers in the linguistic domain to better investigate sentiment expression in sign language. In general, we hope this is an inspiring work for future tight collaborations between linguists and technical practitioners on using technology in the service of a better understanding of sign languages, beyond the widely spread construction of (uninterpretable) translation systems.

\section{Acknowledgements}
This contribution is funded by the German Ministry for Education and Research (BMBF) through the BIGEKO project (grant number 16SV9094), and by the Federal Ministry of Research, Technology and Space (BMFTR) through the RoGSiLT project.

\section{Optional Supplementary Materials: Appendices, Software, and Data}


\subsection{Appendices}
\label{app:all-features}

Tables \ref{tab:features-1-64} and \ref{tab:features-65-96} list the features used to train the best model, sorted by importance.
In the MediaPipe coordinate system: x is the horizontal axis, y is the vertical axis, and z increases with distance from the camera.

\input{features-tab1}
\input{features-tab2}



%
%
%
\bibliographystyle{plain}
\bibliography{LREC2026-SLSentimentAnalysis}



\end{document}

%% file: features-tab1.tex
\begin{table} 
  \centering
  \scriptsize
  \caption{List of the 96 prediction features sorted by importance. Part 1: 1-64.}
    \begin{tabular}{lr}
    \toprule
    \textbf{Feature} & \multicolumn{1}{l}{\textbf{Importance}} \\
    \midrule
    mouthSmileRight\_mean & 0.02707 \\
    mouthSmileLeft\_mean & 0.02354 \\
    browDownLeft\_mean & 0.01867 \\
    pose\_LEFT\_ELBOW\_z\_mean & 0.01661 \\
    mouthUpperUpRight\_mean & 0.01561 \\
    pose\_LEFT\_HIP\_y\_velocity\_std & 0.01530 \\
    pose\_RIGHT\_HIP\_y\_acceleration\_std & 0.01529 \\
    pose\_LEFT\_SHOULDER\_z\_mean & 0.01475 \\
    mouthRight\_mean & 0.01430 \\
    pose\_RIGHT\_HIP\_y\_velocity\_std & 0.01419 \\
    pose\_LEFT\_HIP\_y\_acceleration\_std & 0.01391 \\
    pose\_RIGHT\_ELBOW\_y\_std & 0.01343 \\
    mouthDimpleRight\_std & 0.01322 \\
    mouthDimpleRight\_mean & 0.01305 \\
    jawRight\_mean & 0.01296 \\
    pose\_RIGHT\_HIP\_y\_std & 0.01278 \\
    browDownRight\_mean & 0.01277 \\
    mouthLowerDownLeft\_mean & 0.01266 \\
    mouthSmileRight\_std & 0.01240 \\
    mouthUpperUpLeft\_mean & 0.01211 \\
    pose\_RIGHT\_ELBOW\_y\_mean & 0.01209 \\
    browDownLeft\_std & 0.01190 \\
    pose\_LEFT\_HIP\_y\_std & 0.01184 \\
    head\_pitch\_deg\_mean & 0.01173 \\
    pose\_RIGHT\_HIP\_z\_std & 0.01166 \\
    browOuterUpLeft\_mean & 0.01161 \\
    dist\_elbows\_lr\_avg & 0.01156 \\
    mouthFrownLeft\_std & 0.01133 \\
    browOuterUpLeft\_std & 0.01122 \\
    left\_hand\_WRIST\_y\_mean & 0.01106 \\
    pose\_RIGHT\_HIP\_z\_mean & 0.01102 \\
    eyeLookDownLeft\_mean & 0.01089 \\
    eyeSquintLeft\_mean & 0.01082 \\
    mouthShrugLower\_std & 0.01073 \\
    pose\_RIGHT\_HIP\_y\_mean & 0.01052 \\
    dist\_right\_wrist\_to\_nose\_avg & 0.01044 \\
    head\_yaw\_deg\_mean & 0.01037 \\
    pose\_LEFT\_SHOULDER\_z\_std & 0.01030 \\
    mouthUpperUpRight\_std & 0.01019 \\
    torso\_yaw\_mean & 0.01015 \\
    mouthShrugLower\_mean & 0.01008 \\
    pose\_LEFT\_SHOULDER\_x\_acceleration\_mean & 0.00991 \\
    jawRight\_std & 0.00989 \\
    browOuterUpRight\_mean & 0.00987 \\
    cheekSquintRight\_mean & 0.00984 \\
    mouthRight\_std & 0.00984 \\
    mouthFrownRight\_mean & 0.00973 \\
    right\_hand\_WRIST\_x\_acceleration\_std & 0.00970 \\
    eyeLookDownRight\_mean & 0.00965 \\
    pose\_NOSE\_z\_mean & 0.00963 \\
    R\_WRIST\_accum\_dist\_avg & 0.00934 \\
    pose\_LEFT\_HIP\_z\_std & 0.00932 \\
    pose\_LEFT\_HIP\_z\_velocity\_mean & 0.00926 \\
    mouthRollUpper\_mean & 0.00926 \\
    pose\_LEFT\_ELBOW\_y\_std & 0.00925 \\
    pose\_LEFT\_SHOULDER\_y\_velocity\_std & 0.00917 \\
    right\_hand\_WRIST\_y\_mean & 0.00915 \\
    mouthLowerDownLeft\_std & 0.00914 \\
    browOuterUpRight\_std & 0.00906 \\
    mouthLeft\_std & 0.00897 \\
    browInnerUp\_std & 0.00895 \\
    eyeBlinkLeft\_std & 0.00888 \\
    dist\_right\_wrist\_to\_left\_shoulder\_avg & 0.00883 \\
    mouthPressLeft\_peaks\_per\_s & 0.00879 \\
    \bottomrule
    \end{tabular}%
  \label{tab:features-1-64}%
\end{table}%

%% file: features-tab2.tex
\begin{table}. 
  \centering
  \scriptsize
  \caption{List of the 96 prediction features sorted by importance. Part 2: 65-96.}
    \begin{tabular}{lr}
    \toprule
    \textbf{Feature} & \multicolumn{1}{l}{\textbf{Importance}} \\
    \midrule
    mouthLeft\_peaks\_per\_s & 0.00874 \\
    mouthFrownLeft\_mean & 0.00867 \\
    pose\_RIGHT\_SHOULDER\_z\_mean & 0.00859 \\
    eyeWideRight\_mean & 0.00858 \\
    pose\_RIGHT\_HIP\_z\_velocity\_mean & 0.00858 \\
    eyeBlinkRight\_std & 0.00857 \\
    pose\_LEFT\_HIP\_y\_mean & 0.00846 \\
    noseSneerRight\_std & 0.00846 \\
    pose\_LEFT\_SHOULDER\_y\_acceleration\_mean & 0.00839 \\
    mouthPressRight\_std & 0.00839 \\
    eyeLookInLeft\_peaks\_per\_s & 0.00835 \\
    L\_SHOULDER\_accum\_dist\_avg & 0.00825 \\
    dist\_left\_wrist\_to\_left\_shoulder\_peaks\_per\_s & 0.00824 \\
    pose\_NOSE\_x\_mean & 0.00819 \\
    pose\_RIGHT\_SHOULDER\_z\_velocity\_std & 0.00819 \\
    pose\_RIGHT\_HIP\_z\_acceleration\_std & 0.00816 \\
    pose\_LEFT\_ELBOW\_y\_mean & 0.00810 \\
    torso\_roll\_mean & 0.00807 \\
    mouthPucker\_std & 0.00800 \\
    pose\_RIGHT\_ELBOW\_y\_acceleration\_std & 0.00790 \\
    mouthLowerDownRight\_peaks\_per\_s & 0.00777 \\
    left\_arm\_angle\_mean & 0.00775 \\
    pose\_LEFT\_ELBOW\_x\_mean & 0.00775 \\
    right\_hand\_WRIST\_x\_velocity\_std & 0.00760 \\
    pose\_RIGHT\_SHOULDER\_y\_mean & 0.00751 \\
    left\_hand\_WRIST\_x\_acceleration\_std & 0.00751 \\
    pose\_NOSE\_z\_velocity\_std & 0.00747 \\
    mouthClose\_mean & 0.00734 \\
    pose\_RIGHT\_SHOULDER\_x\_std & 0.00720 \\
    right\_hand\_WRIST\_z\_velocity\_std & 0.00702 \\
    pose\_LEFT\_ELBOW\_y\_peaks\_per\_s & 0.00662 \\
    \bottomrule
    \end{tabular}%
  \label{tab:features-65-96}%
\end{table}%